\documentclass{llncs}
\usepackage{subcaption}
\usepackage{array,float}
\usepackage{amsfonts, amssymb}
\usepackage{graphicx,amsmath}
\usepackage{pgfplots}
\pgfplotsset{compat=1.8}
\usepgfplotslibrary{statistics}
\usepackage{tikz}
\usepackage{hyperref}
\usepackage{caption}
\usepackage{comment}
\usepackage{soul}
\usepackage{tikz-3dplot}
\usepackage[titletoc, title]{appendix}
\usetikzlibrary{%
	calc,%
	decorations.pathmorphing,%
	fadings,%
	shadings%
}
\usetikzlibrary{positioning,arrows}
\tikzstyle{place}=[circle,draw,text width=1.5em, text centered,minimum size=0.1cm]
\usetikzlibrary{arrows.meta}
\usepackage{makeidx}  
\usepackage{multicol}

\DeclareMathOperator{\Tr}{Tr}

\def \Fbf{{\mathbf F}}

\def \Gbf{{\mathbf G}}

\def \Hbf{{\mathbf H}}

\def \Ibf{{\mathbf I}}

\def \Kbf{{\mathbf K}}

\def \mbf{{\mathbf m}}

\def \Pbf{{\mathbf P}}
\def \qbf{{\mathbf q}}
\def \Qbf{{\mathbf Q}}

\def \Rbf{{\mathbf R}}

\def \Sbf{{\mathbf S}}

\def \vbf{{\mathbf v}}

\def \xbf{{\mathbf x}}
\def \Xbf{{\mathbf X}}
\def \ybf{{\mathbf y}}
\def \Ybf{{\mathbf Y}}

\def \0bf{{\mathbf 0}}

\def \Tr{\mbox{Tr}}

\usepackage{lineno}

\begin{document}
\mainmatter              
\title{Extraction of Airways with Probabilistic State-space Models and Bayesian Smoothing}
\titlerunning{EKF}  
%
\author{Raghavendra Selvan\inst{1}, Jens Petersen\inst{1}, Jesper H. Pedersen\inst{2} \and
Marleen de Bruijne\inst{1,3}}
\authorrunning{Raghavendra Selvan et al.} 
%
\tocauthor{Raghavendra Selvan, Jens Petersen, Jesper Pedersen, Marleen de Bruijne}
\institute{Department of Computer Science, University of Copenhagen, Denmark
\and
Department of Cardio-Thoracic Surgery RT, Rigshospitalet, University Hospital of Copenhagen, Denmark
\and
Departments of Medical Informatics and Radiology, Erasmus MC, The Netherlands
\\
\email{raghav@di.ku.dk}\\ 
}
\authorrunning{Raghavendra Selvan et al.} 
%

\maketitle              

\begin{abstract}
Segmenting tree structures is common in several image processing applications. In medical image analysis, reliable segmentations of airways, vessels, neurons and other tree structures can enable important clinical  applications. We present a framework for tracking tree structures comprising of elongated branches using probabilistic state-space models and Bayesian smoothing. Unlike most existing methods that proceed with sequential tracking of branches, we present an exploratory method, that is less sensitive to local anomalies in the data due to acquisition noise and/or interfering structures. The evolution of individual branches is modelled using a process model and the observed data is incorporated into the update step of the Bayesian smoother using a measurement model that is based on a multi-scale blob detector. Bayesian smoothing is performed using the RTS (Rauch-Tung-Striebel) smoother, which provides Gaussian density estimates of branch states at each tracking step. We select likely branch seed points automatically based on the response of the blob detection and track from all such seed points using the RTS smoother. We use covariance of the marginal posterior density estimated for each branch to discriminate false positive and true positive branches. The method is evaluated on 3D chest CT scans to track airways. We show that the presented method results in additional branches compared to a baseline method based on region growing on probability images.

\end{abstract}

\section{Introduction}

Segmentation of tree structures comprising of vessels, neurons, airways etc. are useful in extraction of clinically relevant biomarkers~\cite{vesselreview,exact}. The task of extracting trees, mainly in relation to vessel segmentation, has been studied widely using different methods.  A successful class of these methods are based on techniques from target tracking. Perhaps the most used tracking strategy is to proceed from an initial seed point, make local-model fits to track individual branches  in a sequential manner and perform regular branching checks~\cite{friman,yedidya}. Such methods are prone to local anomalies and can prematurely terminate if occlusions are encountered. The method in~\cite{friman} can overcome such problems to a certain extent using a deterministic multiple hypothesis testing approach; however, it is a semi-automatic method requiring extensive manual intervention and can be computationally expensive. In \cite{yedidya}, vessel tracking on 2D retinal scans is performed using a Kalman filter. They propose an automatic seed point detection strategy using a matched filter. From each of these seed points vessel branches are progressively tracked using measurements that are derived from the image data. A gradient based measurement function is employed which fails in low-contrast regions of the image, which are predominantly regions with thin vessels. Another major class of tracking algorithms are based on a stochastic formulation of tracking~\cite{florin,lesage2016adaptive} using some variation of particle filtering. Particle filter-based methods are known to scale poorly with dimensions of the state space~\cite{vesselreview}. 

In spirit, we propose an exploratory method like particle filter-based methods, with a salient distinction that the proposed method can track branches from several seed points across the volume. We use linear Bayesian smoothing to estimate branch states, described using Gaussian densities. Thus, the method inherently provides an uncertainty measure, which we use to discriminate true and false positive branches. Further, unlike particle filter-based methods, the proposed method is fast, as Bayesian smoothing is implemented using the RTS (Rauch-Tung-Striebel) smoother~\cite{simo} involving only a set of linear equations.

\section{Method}
\label{sec:method}
We formulate tracking of branches in tree structures using probabilistic state-space models, commonly used in target tracking and control theory~\cite{simo}. The proposed method takes image data as input and outputs a collection of disconnected branches that taken together forms the tree structure of interest. We first process the image data to obtain a sequence of measurements and track all possible branches individually using Bayesian smoothing. We then use covariance estimates of individual branches to output a subset of the most likely branches yielding the tree structure of interest. Details of this process are described below.

\subsection{Tracking individual branches}

We assume the tree structure of interest, $\Xbf$, to be a collection of $T$ independent random variables $\Xbf = \{\Xbf_1, \Xbf_2, \dots, \Xbf_T\}$, where individual branches are denoted $\Xbf_i$. Each branch $\Xbf_i$ of length $L_i$ is treated as a sequence of states, $\Xbf_i = [\xbf_0, \xbf_1, \dots, \xbf_{L_i}]$. These states are assumed to obey a first-order Markov assumption, i.e.,
\begin{equation}
p(\xbf_k| \xbf_{k-1}, \xbf_{k-2}, \dots, \xbf_{0}) = p(\xbf_k | \xbf_{k-1}).
\label{eq:markov}
\end{equation}
The state vector has seven random variables,
\begin{equation}
\xbf_k = [x, y, z, r, v_x, v_y, v_z]^T,
\end{equation}
describing a tubular segment centered at Euclidean coordinates $[x,y,z]$, along an axis given by the direction vector $[v_x, v_y, v_z]$ with radius $r$.

The observed data, image $\Ibf$, is processed to be available as a sequence of vectors. We model the measurements as four dimensional state vectors consisting only of position and radius. This is accomplished using a multi-scale blob detector~\cite{lindberg}. The input image $\Ibf$  with $N_v$ voxels is transformed into a sequence of $N$ measurements, with position and radius information, denoted $\Ybf = [\ybf_0, \dots, \ybf_N]$, where each $\ybf_i = [x, y, z, r]^T$. This procedure applied to the application of tracking airway trees is described in Section~\ref{subsec:ms}.

\begin{figure}[t]
	\centering
	\includegraphics[width = 0.7\textwidth]{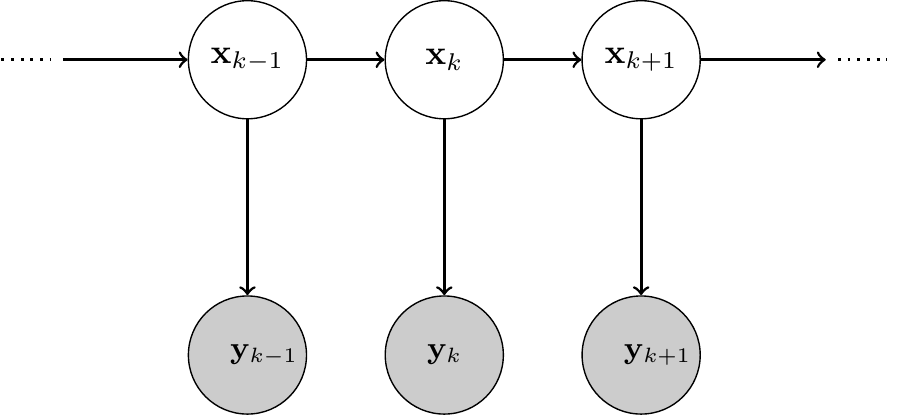}
	\caption{Bayesian network view of the relation between the underlying true states, $\xbf_i$, and the measurements, $\ybf_i$, for a single branch. }
	\label{fig:bayesNet}
\end{figure}

\subsection{Process and Measurement Models}
\label{subsec:model}

Transition from one tracking step to another within a branch is modelled using the process model. We use a process model that captures our understanding of how individual branches evolve between tracking steps and has similarities with the model used in~\cite{yedidya}. We assume first-order Markov independence in state transitions from~\eqref{eq:markov}, captured in the process model below:
\begin{equation}
\xbf_k = \Fbf \xbf_{k-1} + \qbf = \begin{bmatrix}
1 & 0 & 0 & 0 & \Delta & 0 & 0 \\
0 & 1 & 0 & 0 & 0 & \Delta & 0 \\
0 & 0 & 1 & 0 & 0 & 0 & \Delta \\
0 & 0 & 0 & 1 & 0 & 0 & 0 \\
0 & 0 & 0 & 0 & 1 & 0 & 0 \\
0 & 0 & 0 & 0 & 0 & 1 & 0 \\
0 & 0 & 0 & 0 & 0 & 0 & 1
\end{bmatrix} \begin{bmatrix}
x_{k-1} \\ y_{k-1} \\ z_{k-1} \\ r_{k-1} \\ {v_x}_{k-1} \\ {v_y}_{k-1} \\ {v_z}_{k-1}
\end{bmatrix} + \qbf
\label{eq:motMod}
\end{equation}
where $\mathbf{F}$ is the process model function and $\qbf$ is the process noise. $\qbf$ is assumed to be a zero mean Gaussian density, i.e, $\qbf \sim N(\mathbf{0}, \Qbf)$, with process covariance, $\Qbf_{7\times7}$,  acting only on direction and radius components of the state vector,
\begin{equation}
	{\Qbf}_{[4:7,4:7]} =  \sigma_q^2 \Delta \times  \Ibf_{4\times4},
\end{equation}
where only the non-zero part of the matrix is shown and $\sigma_q^2$ is the process variance. The parameter $\Delta$ can be seen as step size between tracking steps. 
As~\eqref{eq:motMod} is a recursion, the initial point (seed point), $\xbf_0$, comprising of position, scale and orientation information is provided to the model. Seed points are assumed to be described by Gaussian densities, $\xbf_0 \sim N(\hat{\xbf_0}, \Pbf_0)$, with mean $\hat{\xbf_0}$ and covariance $\Pbf_0$. We present an automatic strategy to detect such initial seed points in~\ref{subsec:seed}.

The measurement model describes the relation between each of the 4-D measurements, $\ybf_k$ in the sequence, $\Ybf = [\ybf_1,\dots,\ybf_N]$, and the state vector, $\xbf_k$, as shown in Figure~\ref{fig:bayesNet}. A simple linear measurement model captures this relation,
\begin{equation}
\ybf_{k} = \Hbf \xbf_k + \mbf = \begin{bmatrix}
1 & 0 & 0 & 0 \\
0 & 1 & 0 & 0 \\
0 & 0 & 1 & 0 \\ 
0 & 0 & 0 & 1 \\
0 & 0 & 0 & 0 \\ 
0 & 0 & 0 & 0 \\
0 & 0 & 0 & 0
\end{bmatrix} \begin{bmatrix} 
x_{k} \\ y_{k} \\ z_{k} \\ r_{k} \\ v_{{x}_{k}} \\ v_{{y}_{k}} \\ v_{{z}_{k}}\end{bmatrix} + \mbf	
\label{eq:meas}
\end{equation}
where $\ybf_k$ are observations generated by true states of the underlying branch at step $k$, $\Hbf$ is the measurement function. $\mbf \sim N(\mathbf{0},\Rbf)$ is the measurement noise with covariance $\Rbf$ that is a diagonal matrix with entries, $[\sigma^2_{m_x},\sigma^2_{m_y},\sigma^2_{m_z},\sigma^2_{m_r}]$, which correspond to variance in the observed position and radius, respectively. All possible measurement vectors obtained from the image are aggregated into the measurement variable $\Ybf$.

\subsection{Bayesian Smoothing}
\label{sec:ekf}

The state-space models presented above enable us to estimate branches using the posterior distributions, $p(\Xbf_i|\Ybf) \forall i = [0,\dots,T]$, using standard Bayesian methods. We employ Bayesian smoothing as all the measurements are available at once, when compared to sequential observations that are more common in object tracking applications. Due to a linear, Gaussian process and measurement models, Bayesian smoothing can be optimally performed using the RTS smoother~\cite{simo}. RTS smoother uses two Bayesian filters to perform forward filtering and backward smoothing. Forward filtering is identical to performing Kalman filtering and consists of sequential prediction and update with observed information of the state variable. Once a branch is estimated using forward filtering, the saved states are used to perform backward smoothing using a Kalman-like filter which improves state estimates by incorporating additional information from future steps. Standard equations for an RTS smoother are presented below~\cite{simo}.
\begin{table}
\caption{Standard RTS Smoother Equations}
\begin{multicols}{2}
\centering
\textbf{Forward Filtering}
{
\begin{align}
	\label{eq:meanPred} \hat{\xbf}_{k|k-1} &= \Fbf\hat{\xbf}_{k-1|k-1} \\
	\label{eq:covPred} \Pbf_{k|k-1} &= \Fbf\Pbf_{k-1|k-1}\Fbf^T + \Qbf \\
	\label{eq:predMeas} \vbf_k & = \ybf_k - \Hbf\hat{\xbf}_{k|k-1} \\
	\label{eq:predCov} \Sbf_k & = \Hbf\Pbf_{k|k-1}\Hbf^T + \Rbf \\
	\label{eq:gain} \Kbf_k & =  \Pbf_{k|k-1}\Hbf^T\Sbf_k^{-1} \\
	\label{eq:postMean} \hat{\xbf}_{k|k} & = \hat{\xbf}_{k|k-1} + \Kbf_k\vbf_k \\
	\label{eq:postCov} \Pbf_{k|k} & =  \Pbf_{k|k-1} - \Kbf_k\Sbf_k\Kbf_{k}^T
\end{align}
}
\textbf{Backward Smoothing}
{
\begin{align}
	\label{eq:bgain} \Gbf_k & =  \Pbf_{k|k}\Fbf^T\Pbf_{k+1|k}^{-1} \\ 
	\label{eq:smoMean} \hat{\xbf}_{k|L} & = \hat{\xbf}_{k|k} + \Gbf_k(\hat{\xbf}_{k+1|L} - \hat{\xbf}_{k+1|k}) \\
	\label{eq:smoCov} \Pbf_{k|L} & =  \Pbf_{k|k} - \Gbf_k(\Pbf_{k+1|k} - \Pbf_{k+1|L})\Gbf^T 
\end{align}
}
\end{multicols}
\vspace{-1cm}
\label{tab:rts}
\end{table}

\subsubsection{Forward Filtering} Equations in the first column of Table~\ref{tab:rts} are used to perform prediction and update steps of the forward filtering. In the prediction step, process model is used to predict states at the next step. Mean $\hat{\xbf}_{k|k-1}$ and covariance $\Pbf_{k|k-1}$ estimates of the predicted Gaussian density, i.e, of state $k$ conditioned on the previous state, denoted with subscript $k|k-1$, are computed in ~\eqref{eq:meanPred},\eqref{eq:covPred}.  In the update step, described in~\eqref{eq:predMeas} -- ~\eqref{eq:postCov}, predicted density is associated with a measurement vector to obtain posterior density. First, the new information from measurement $\ybf_k$ is computed using~\eqref{eq:predMeas} and is aptly called the ``innovation", denoted as $\vbf_k$. Uncertainty in the new information, innovation covariance $\Sbf_k$, is computed in~\eqref{eq:predCov}. Then, predicted mean is adjusted with weighted innovation and predicted covariance is adjusted with weighted innovation covariance to obtain the posterior mean and covariances, in~\eqref{eq:postMean} and~\eqref{eq:postCov}, respectively. The weighting computed in~\eqref{eq:gain}, denoted as $\Kbf_k$, is the Kalman gain which controls the extent of information fusion from process and measurement models. 

We continue estimation of the posterior density (described by posterior mean and covariance) in a sequential manner for the branch until no new measurements exist for updating. After the final update step, a sequence of posterior mean estimates $[\hat{\xbf}_{0|0}, \dots, \hat{\xbf}_{L_i|L_i}]$ and posterior covariance estimates $[\Pbf_{0|0}, \dots, \Pbf_{L_i|L_i}]$, obtained from the forward filter are saved, for further use by the backward smoother. 

\subsubsection{Backward smoothing} 

The smoothed estimates are obtained by running a backward filter starting from the final tracked state of the forward filter. The intuition behind backward smoothing is that the uncertainty in making predictions in the forward filtering can be alleviated using information from future steps. It is implemented using the equations in the second column of Table~\ref{tab:rts}. 
\subsubsection{Gating}
\label{subsec:gating}

When performing the RTS smoother recursions, the forward filter expects a single measurement vector for the update step. We employ rectangular and ellipsoidal gating to reduce the number of measurements handled during the update step~\cite{bar2011tracking}. 

First, we perform simple rectangular gating which is based on excluding measurements that are outside a rectangular region around the predicted measurement $\Hbf\hat{\xbf}_{k|k-1}$ in equation~\eqref{eq:predMeas} using the following condition:
\begin{equation}
	|\ybf_i - \Hbf\xbf_{k|k-1}| \leq \kappa \times \text{diag}(\Sbf_k), \forall \ybf_i \in \Ybf
\end{equation}
where $\Sbf_k$ is the covariance of the predicted measurement in equation~\eqref{eq:predCov}. The rectangular gating coefficient, $\kappa$, is usually set to a value $\geq 3$~\cite{bar2011tracking}. Rectangular gating localises the number of candidate measurements relevant to the current tracking step. To further narrow down on the best candidate measurement for update, we follow rectangular gating with ellipsoidal gating~\cite{bar2011tracking}. With ellipsoidal gating we accept the measurements within the ellipsoidal region of the predicated covariance, using the following rule:
\begin{equation}
	(\Hbf\xbf_{k|k-1} - \ybf_i)^T \Sbf_k^{-1}(\Hbf\xbf_{k|k-1} - \ybf_i) \leq G
	\label{eq:ell}
\end{equation}
where $G$ is the rectangular gating threshold, obtained from the gating probability  $P_g$, which is the probability of observing the measurement within the ellipsoidal gate,
\begin{equation}
	P_g = 1- \exp \Big (-\frac{G}{2}\Big ).
\end{equation}

\subsection{Tree as a Collection of Branches}

Once a branch is smoothed and saved using Bayesian smoothing described previously, we process new seed points and start tracking branches until no further seed points remain to track from. This procedure yields a collection of disconnected branches. The next task is to obtain a subset of likely branches that represent the tree structure of interest by discarding false positive branches. 

\subsubsection{Validation of Tracked Branches}
\label{subsec:valid}

An advantage of using Bayesian smoothing to track individual branches is that apart from estimating the branch states from the image data (using the smoothed posterior mean estimates), we can also quantify the uncertainty of the estimation at each tracking step (using the smoothed posterior covariance estimates). Thus, we have the possibility of aggregating this uncertainty over the entire branch to validate them. We explore this notion to create a criterion for accepting or rejecting branches.

By aggregating variance for all tracking steps in each branch, we obtain a measure of the quality of branches. A straightforward approach is to use total variance, obtained using the trace of each of the smoothed posterior covariance matrices. We average the sum total variance over the length of each branch, $l_i$, to obtain a score, $\mu_i$, which is then thresholded by  a cut-off $\mu_c$ to qualify the branches,
\begin{equation}
	\mu_i = \frac{\sum_{k=1}^{l_i} \Tr(\Pbf_{k|k})}{l_i}.
			\label{eq:valid}
\end{equation}

\subsection{Application to Airways}
\label{subsec:airway}

The proposed method for tracking tree structures can be applied to track airways, vessels or other tree structures encountered in image processing applications. We focus on tracking airways from lung CT data and present the specific strategies used to implement the proposed method.

\begin{figure}[t]
\begin{center}
\begin{subfigure}{0.3\textwidth}
\centering
\includegraphics[width=0.94\textwidth]{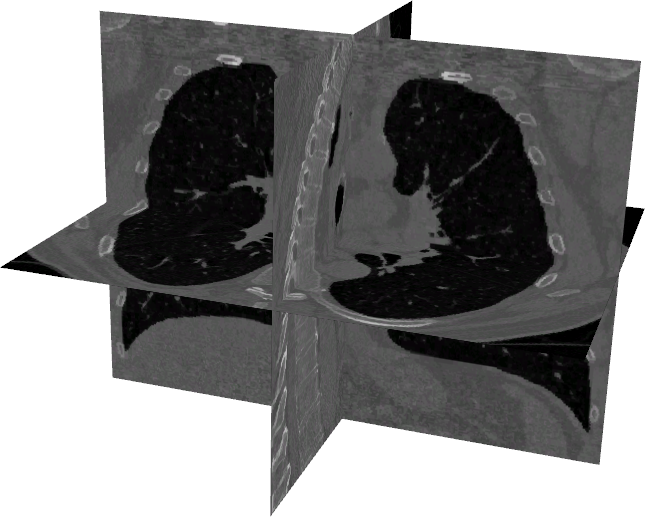}
\caption{Intensity image}
\end{subfigure}
\begin{subfigure}{0.3\textwidth}
\centering
\includegraphics[width=0.9\textwidth]{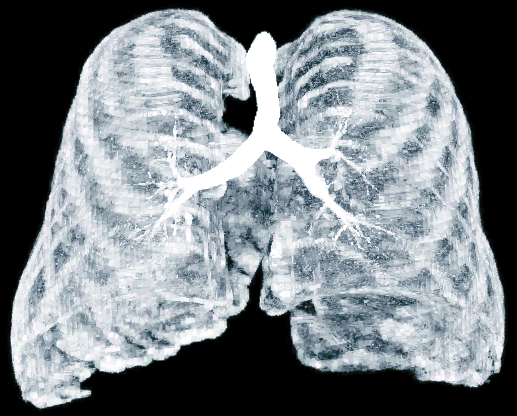}
\caption{Probability image}
\end{subfigure}
\begin{subfigure}{0.38\textwidth}
\centering
\includegraphics[width=0.9\textwidth]{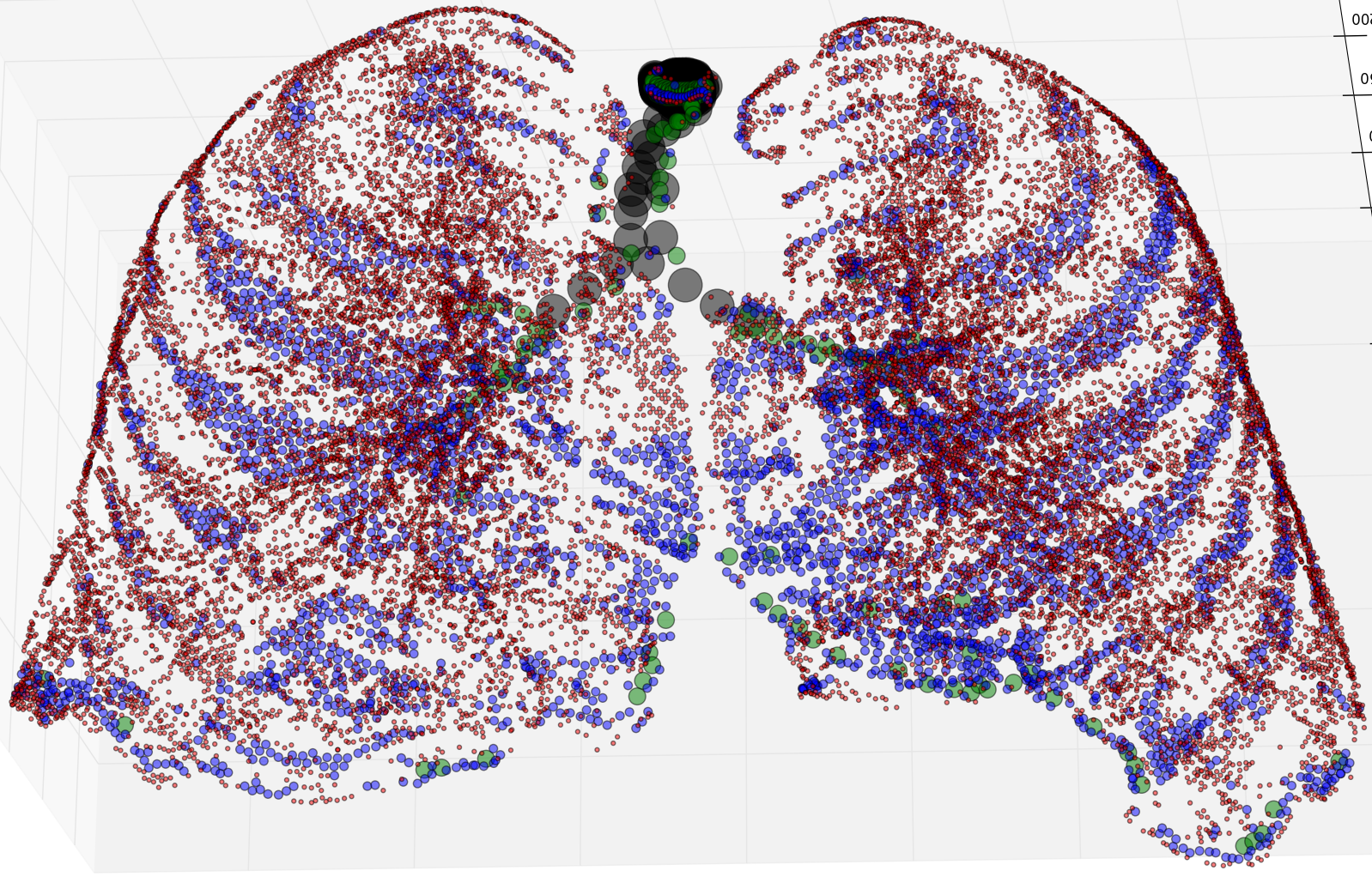}
\caption{Multi-scale blob image}
\end{subfigure}
\caption{The pipeline of image representations, ultimately showing the multi-scale representation.}
\label{fig:ms}
\end{center}
\vspace{-1cm}
\end{figure}
\subsubsection{Multi-scale representation} 
\label{subsec:ms}
The measurement model discussed in Section~\ref{subsec:model} assumes a 4-D state vector as measurements to the RTS smoother. This is achieved by first computing an airway  probability image using a k-Nearest Neighbour voxel classifier trained to discriminate between airway and background, described in~\cite{vessel}. Blob detection with automatic scale selection~\cite{lindberg} for different scales, $\sigma_s = (1,2,4,8,12)mm$, is performed on the probability image to obtain the 4D state measurements as blob position and radius. Indistinct blobs are removed if the absolute value of the normalized response at the selected scale, $\sigma_s^*$, is less than a threshold~\cite{lindberg}. This makes the representation sparse, $N << N_v$, and the tracking more efficient than if performed at voxel-level. An example of the sparse representation can be found in Figure~\ref{fig:ms}. 

%
\vspace{-0.2cm}
\subsubsection{Initialisation of Branches} 
\label{subsec:seed}
The multi-scale representation of the image data discussed above also provides a response corresponding to the best scale. As this response is normalised for scales, we incorporate this information in selecting the initial seed point for every branch. We start tracking from the seed point with the largest scale and the largest response. The initial direction information is obtained from eigen value analysis of the Hessian matrix computed at the corresponding scale provided in the measurement vector. Once a branch is tracked along the initial direction, we track from the same seed point but in the opposite direction. Thus, if a seed point is obtained from the middle of a branch we can track it bidirectionally. After tracking in both directions, all the involved measurements including the seed point are removed from the measurement vector, and the next best candidate seed point is chosen and tracking commences  from there. The tracking procedure on the entire image is complete when no more seed points are available. 

\section{Experiments and Results}
\label{sec:res}

\subsection{Data}
The evaluation was carried out on 32 low-dose CT chest scans from a lung cancer screening trial~\cite{dlcst}. Training and test sets comprising of 16 images each were randomly obtained from the data set. All scans have a resolution of approximately 1mm $\times$ 0.78mm $\times$ 0.78mm. The reference segmentations consist of expert verified union over the results of two previous methods~\cite{vessel,lop}. The proposed method is compared with region growing on the probability images.

\subsection{Error Measure, Initial Parameters and Tuning}
We use an error measure defined as the average of two distances, $d_{err} = (d_{FP} + d_{FN})/2 $. The first distance, $d_{FP}$, captures the false positive error and is the average minimum Euclidean distance from segmentation centerline points to reference centerline points. $d_{FN}$ similarly defines the false negative error, as the average minimum Euclidean distance from reference centerlines points to segmentation centerline points.

There are several parameters related to the RTS smoother that need to be initialised. These parameters were tuned using the training set and fixed for the evaluation on the test set to: standard deviations of the process noise, $\sigma_q = 0.3$, measurement noise on radius $\sigma_{m_r} = 1$ mm and measurement noise on position $(\sigma_{m_x}, \sigma_{m_y}, \sigma_{m_z}) = 2$ mm. The initial covariance, $\Pbf_0$ across branches was set to $\Ibf_{7\times7}$. The most crucial parameter in the proposed method is the threshold parameter $\mu_c$ presented in Section~\ref{subsec:valid}. The threshold to validate branches is tuned to be $\mu_c = 2.0$. The gating probability was set to a high value, $P_g = 0.99$~\cite{bar2011tracking}.

\subsection{Results}

\begin{figure}[t!]
\begin{center}
\begin{subfigure}{0.3\textwidth}
\centering
\includegraphics[width=1.0\textwidth]{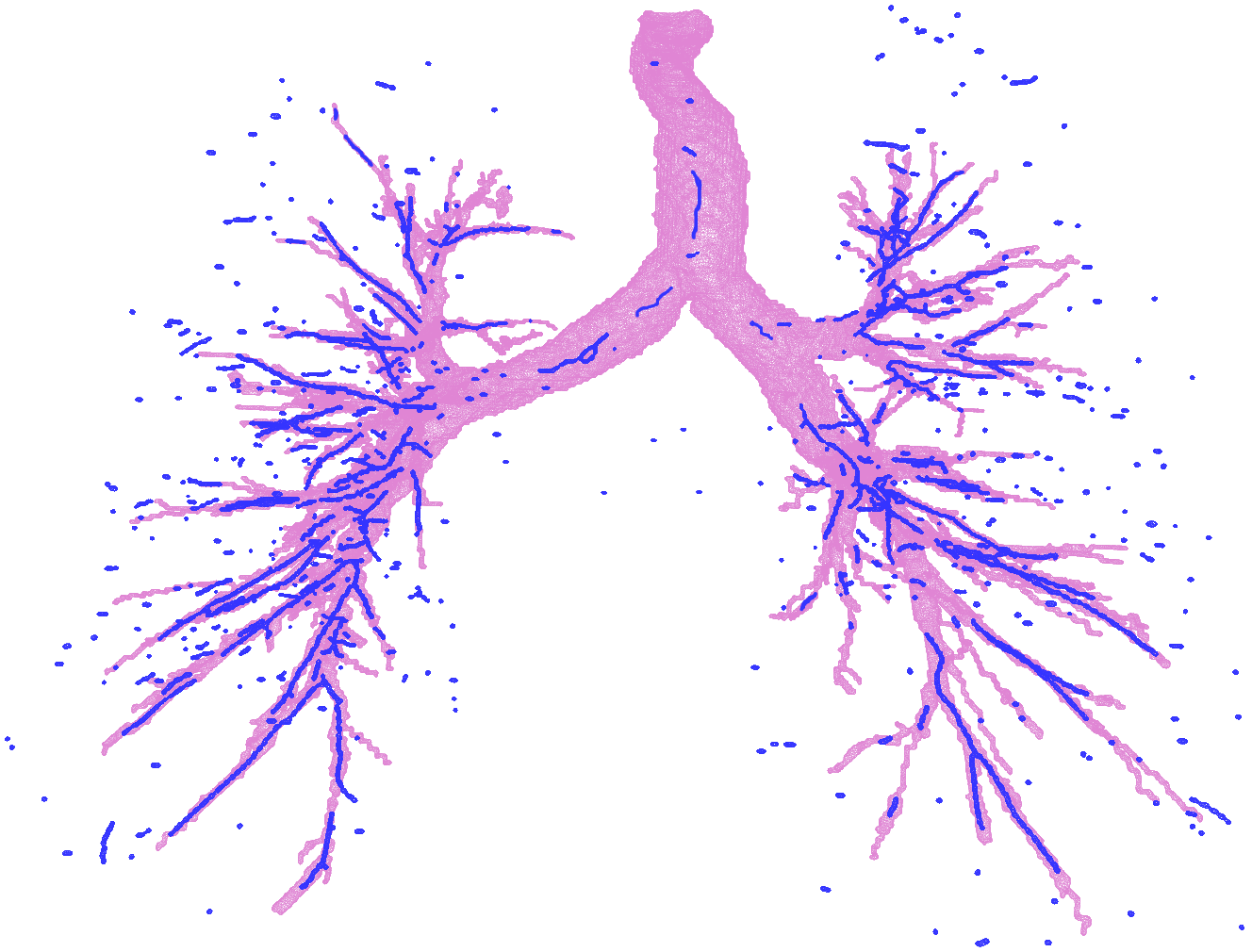}
\caption{}
\label{fig:noThresh}
\end{subfigure}
\begin{subfigure}{0.28\textwidth}
\centering
\includegraphics[width=1.0\textwidth]{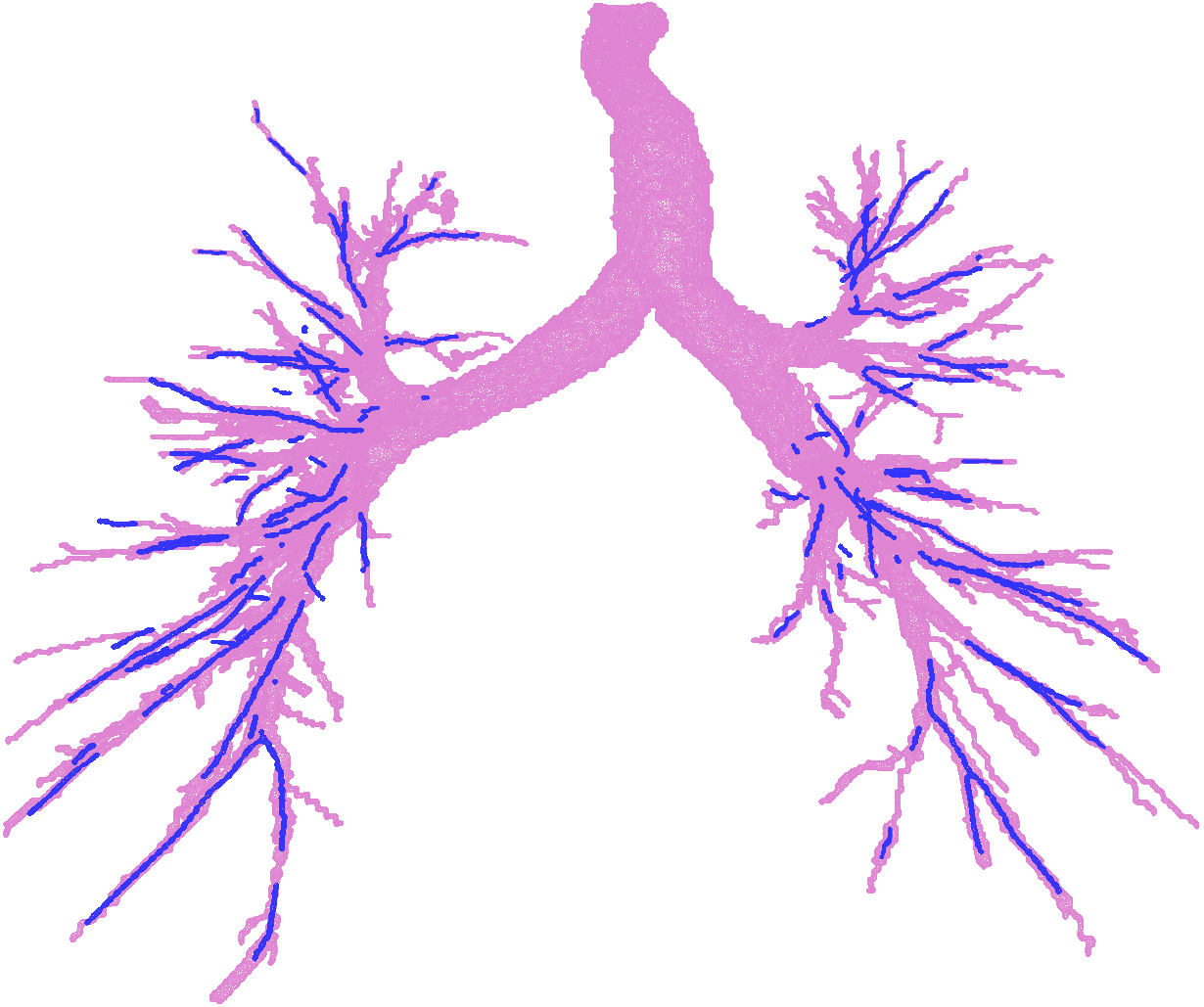}
\caption{}
\label{fig:thresh}
\end{subfigure}
\begin{subfigure}{0.28\textwidth}
\centering
\includegraphics[width=1.0\textwidth]{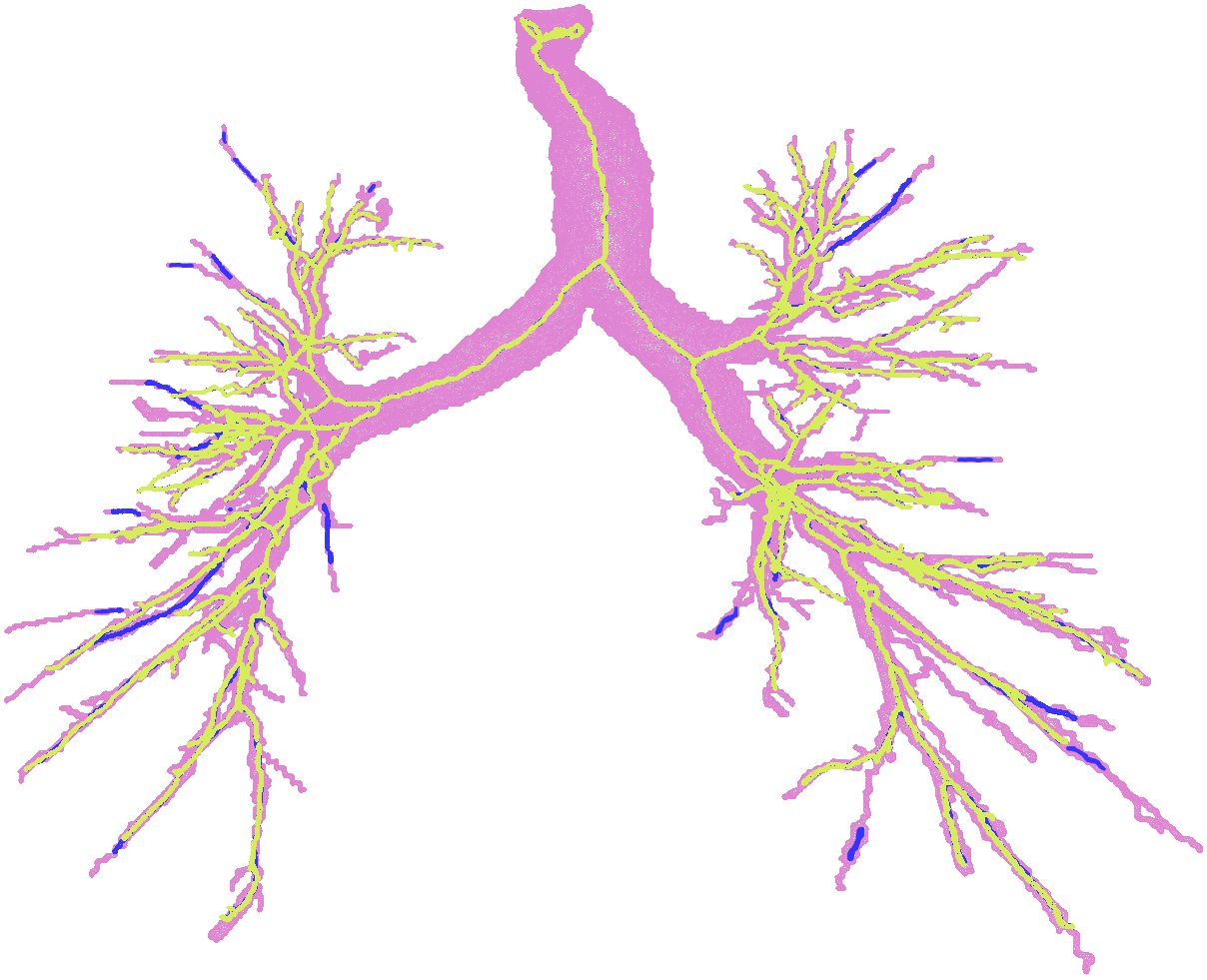}
\caption{}
\label{fig:comb}
\end{subfigure}
\caption{Visualisation of the centerlines extracted using the proposed method before and after thresholding to discard false positive branches overlaid on the reference segmentation, shown in (a) and (b) respectively. The combined results from the proposed method and region growing on probability is shown as the blue centerline in (c).}
\label{fig:vis}
\vspace{-1cm}
\end{center}
\end{figure}

Figure~\ref{fig:vis} illustrates features of the proposed method by visualising centerlines overlaid on the reference segmentation. Influence of the threshold parameter $\mu_c$ is illustrated with the segmentation results for a single volume without any threshold (seen in Figure~\ref{fig:noThresh}) and after applying the tuned threshold (seen in Figure~\ref{fig:thresh}). Evidently, thresholding the average total variance of a branch eliminates false positive branches.

The final output obtained from the method is a collection of disconnected branches. While such collection of branches are still useful in extracting biomarkers, for evaluation purposes we merge the results obtained with the segmentations from region growing on probability images and extract centerlines from the merged segmentation using 3D thinning,  as seen in Figure~\ref{fig:noThresh} and ~\ref{fig:thresh}. This also allows us to demonstrate the improvement our method provides by extracting peripheral airway branches, which are typically the challenging ones. One such combined result is shown in Figure~\ref{fig:comb}, where the yellow centerlines correspond to region growing and blue one is the combined result. 

\newcolumntype{K}[1]{>{\centering\arraybackslash}p{#1}}
\vspace{-0.75cm}
\begin{table*}[h]
\caption{ Performance comparison on the test set}
\label{tab:res}
\begin{center}
\small{
    \begin{tabular}{| l | K{2cm} | K{1.75cm}|  K{1.75cm}| K{2.25cm}|}
    \hline
    Method & $d_{FP}$(mm) & $d_{FN}$(mm) & $d_{err}$ (mm) & Std.Dev. (mm)\\ \hline
    RG & $0.423$ & $3.579$ & $2.001$ & $0.208$ \\ \hline
    (RTS+RG)$_1$ & $0.449$ & $2.102$ & $1.276$ & $0.187$  \\ \hline
    (RTS+RG)$_2$ & $0.401$ & $2.658$ & $1.529$ & $0.165$  \\ \hline
    \end{tabular}}
\end{center}
\end{table*}
\vspace{-0.75cm}

Performance on the test set for two different scenarios of the proposed method is reported in Table~\ref{tab:res} along with the numbers for region growing on probability images. The result for the best performing region growing on probability images is denoted with RG and those obtained by combining the proposed method with region growing are denoted as RTS+RG. We first combine the proposed method with the best performing region growing case (with minimum $d_{err}$) results and it is denoted as (RG+RTS)$_1$. We observe an improvement of about 36\% on $d_{err}$. It is to be noted, there is substantial reduction in $d_{FN}$, indicating that many branches missed by region growing are now segmented. There is a very small increase in false positives which could also be due to the missing branches in the reference segmentation; however, the net result is a large improvement. To test whether the proposed method can simultaneously reduce the number of false positives and false negatives compared to region growing, we merge the proposed method with the region growing result that yields non-optimal $d_{err}$, and do observe a reduction in both $d_{FP}$ and $d_{FN}$ when compared to the best performing RG as seen in the entries for (RG+RTS)$_2$.

The computational expense for running the proposed method is small. The largest chunk of it is used in generating the multi-scale representation of the images, which is in the range of 10-15s per volume. Tracking using the RTS smoother and obtaining the segmentation takes about 4s on a laptop with 8 cores and 32 GB memory running Debian operating system.
\vspace{-0.25cm}
\section{Discussion and Conclusions}
\label{sec:disc}
We presented an automatic method for tracking tree structures, in particular airways, using probabilistic state-space models and Bayesian smoothing. We demonstrated that branches can be tracked individually from across the volume, starting from several seed points. This approach of tracking branches from across the volume has the advantage that even in the presence of occlusions, such as mucous plugging or image acquisition noise, the chances of detecting branches beyond the occlusions are higher. An inherent measure of uncertainty in the branch estimates has been presented due to the Bayesian nature of the method. We demonstrated the use of thresholding this uncertainty measure to discriminate detected branches. The use of sparse representation of voxels in the image using blob detection makes the method computationally efficient.

A possible limitation with the proposed method is that it yields a disconnected tree structure. For applications where this is an issue, one can enforce a global connectivity constraint on the disconnected set of branches to obtain fully connected tree as done in~\cite{connect} or similar. It is also possible to derive biomarkers directly from the disconnected branches, as shown in~\cite{treeShape}.

We performed an evaluation of the results obtained from the proposed method by combining it with the results from region growing on probability images. We showed that there is substantial improvement in the segmentation results, indicating that the exploratory approach taken up in our method has potential in improving tree segmentations.
\vspace{-0.25cm}
\section{Acknowledgements}

This  work was funded by the Independent Research Fund Denmark (DFF) and Netherlands Organisation for Scientific Research (NWO).

\end{document}